%% file: main.tex
\documentclass[10pt,twocolumn,letterpaper]{article}

\usepackage{iccv}      %

\usepackage{graphicx}
\usepackage{amsmath}
\usepackage{amssymb}
\usepackage{booktabs}
\usepackage{caption}
\usepackage{comment}
\usepackage{cite}
\usepackage{amsmath,amssymb} %
\usepackage{color}
\usepackage{varwidth}
\usepackage{xcolor}
\usepackage{lipsum}  
\usepackage{verbatim}
\usepackage{multirow}
\usepackage{adjustbox}
\usepackage{array}
\usepackage{bbm}
\usepackage{mathtools}
\usepackage{pifont}
\usepackage{scalerel}
\usepackage{listings}
\newcolumntype{x}[1]{>{\centering\let\newline\\\arraybackslash\hspace{0pt}}p{#1}}

\input{math_commands.tex}

\definecolor{iccvblue}{rgb}{0.21,0.49,0.74}
\usepackage[pagebackref,breaklinks,colorlinks,allcolors=iccvblue]{hyperref}
\usepackage{url}

\def\paperID{1} %
\def\confName{ICCV}
\def\confYear{2025}

\title{Imagining the Unseen: Generative Location Modeling for Object Placement}

\author{
Jooyeol Yun$^{1*}$, Davide Abati$^{1}$, Mohamed Omran$^{1}$, Jaegul Choo$^{2}$, Amirhossein Habibian$^{1}$, Auke Wiggers$^{1}$\\
$^{1}$ Qualcomm AI Research$^{\dagger}$, $^{2}$ Korea Advanced Institute of Science \& Technology\\
\small{\texttt{\{blizzard072,jchoo\}@kaist.ac.kr}}\\
\small{\texttt{\{dabati,momran,ahabibia,auke\}@qti.qualcomm.com}}
}

\definecolor{updatered}{RGB}{235, 64, 52}

\definecolor[named]{ACMDarkBlue}{cmyk}{1,0.58,0,0.21}
\hypersetup{colorlinks,
  linkcolor=ACMDarkBlue,  %
  citecolor=ACMDarkBlue,  %
  urlcolor=ACMDarkBlue,   %
  filecolor=ACMDarkBlue}
\renewcommand*\backref[1]{\ifx#1\relax \else #1 \fi}
  
\newcommand\blfootnote[1]{%
  \begingroup
  \renewcommand\thefootnote{}\footnote{#1}%
  \addtocounter{footnote}{-1}%
  \endgroup
}

\begin{document}
\maketitle
\blfootnote{$^*$Work completed during an internship at Qualcomm Technologies, Inc.}
\blfootnote{$^{\dagger}$Qualcomm AI Research is an initiative of Qualcomm Technologies, Inc.}
\input{sections/0_abstract}
\input{sections/1_introduction}
\input{sections/2_background}
\input{sections/3_method}

\input{sections/4_experiments}

\input{sections/5_discussion}
\input{sections/6_conclusion}
\clearpage
{
    \small
    \bibliographystyle{ieeenat_fullname}
    \bibliography{main}
}
\clearpage
\input{sections/7_appendix}

\end{document}

%% file: math_commands.tex
\usepackage{amsmath,amsfonts,bm}

\def\Secref#1{Section~\ref{#1}}

\def\eqref#1{equation~\ref{#1}}
\def\Eqref#1{Equation~\ref{#1}}

\def\1{\bm{1}}

\DeclareMathAlphabet{\mathsfit}{\encodingdefault}{\sfdefault}{m}{sl}
\SetMathAlphabet{\mathsfit}{bold}{\encodingdefault}{\sfdefault}{bx}{n}

\newcommand{\E}{\mathbb{E}}
\newcommand{\Ls}{\mathcal{L}}

%% file: sections/0_abstract.tex
\begin{abstract}

Location modeling, or determining where \emph{non-existing} objects could feasibly appear in a scene, has the potential to benefit numerous computer vision tasks, from automatic object insertion to scene creation in virtual reality. 
Yet, this capability remains largely unexplored to date.
In this paper, we develop a \emph{generative} location model that, given an object class and an image, learns to predict plausible bounding boxes for such an object. 
Our approach first tokenizes the image and target object class, then decodes bounding box coordinates through an autoregressive transformer. 
This formulation effectively addresses two core challenges in locatio modeling: the inherent one-to-many nature of plausible locations, and the sparsity of existing location modeling datasets, where fewer than 1\% of valid placements are labeled. 
Furthermore, we incorporate Direct Preference Optimization to leverage negative labels, refining the spatial predictions.
Empirical evaluations reveal that our generative location model achieves superior placement accuracy on the OPA dataset as compared to discriminative baselines and image composition approaches.
We further test our model in the context of object insertion, where it proposes locations for an off-the-shelf inpainting model to render objects.
In this respect, our proposal exhibits improved visual coherence relative to state-of-the-art instruction-tuned editing methods, demonstrating a high-performing location model's utility in a downstream application.  
\end{abstract}

%% file: sections/1_introduction.tex
\section{Introduction}
Identifying plausible locations for objects that do not exist in an image is an underexplored problem in computer vision. 
Traditional perception methods, such as object detectors~\citep{detr,fasterrcnn}, primarily focus on detecting and localizing existing objects within scenes. %
However, numerous emerging applications demand a complementary capability, which is to  determine where \emph{non-existing objects could realistically appear}. 
For example, object insertion~\citep{powerpaint,gligen,paintbyexample}, generative data augmentation~\citep{xpaste, kupyn2024dataset, fang2024dataaugmentation}, content planning in virtual reality~\citep{park2005virtualrealityplacement}, and robotics scenarios~\citep{cheong2020whererelocaterobotics} require models to reason about objects and their spatial relations, even if some of them are not yet present in the scene.

\input{floats/figures/intro_example}
\footnotetext{OPA images are under the MIT License: \url{https://github.com/bcmi/Object-Placement-Assessment-Dataset-OPA/blob/main/LICENSE}}

One can make this task explicit by training a \emph{location model}.
Training such a model poses significant challenges, as the task is inherently ambiguous and suffers from data scarcity. 
Unlike object detection, where existing objects are precisely annotated, annotations with plausible insertion locations typically lack exhaustive labels. 
Even, specialized datasets explicitly created for object placement~\citep{opa, pbi} annotate fewer than 1\% of possible insertion points, as it is impractical to exhaustively label every region in the image.
Previous approaches have often addressed this by either treating unlabeled regions as implicitly invalid~\citep{lin2018st, tripathi2019learning, placenet} or relying on custom loss functions that heavily penalize unlabeled but potentially valid placements~\citep{fopa, topnet}. 
Such methods are often sensitive to specific annotation distributions within datasets (\eg, number of annotation per image), hindering their generalization to newly released datasets.

In this paper, we overcome these limitations by recasting the problem from a \emph{generative} perspective. 
Generative models naturally address data sparsity by directly modeling the distribution of plausible placements, enabling effective training from sparse positive examples without requiring exhaustive annotations.
Specifically, we encode the input image along with the intended object class into a sequence of tokens, and train an autoregressive transformer~\citep{transformers, gpt2} to iteratively generate plausible bounding-box coordinates. 
Additionally, when negative annotations (indicating unsuitable locations) are available, we incorporate these as explicit preferences, favoring positive placements over negative ones through Direct Preference Optimization~\citep{dpo}, which further improves the model's accuracy.

Our experiments demonstrate that our location model consistently predicts more accurate and plausible object locations compared to existing discriminative methods, highlighting the advantages of a generative formulation.
Some exemplar locations are illustrated in \cref{fig:intro-example}.
To further showcase the practical utility of our model, we conduct comprehensive experiments in an \emph{object insertion} scenario, where the predicted locations guide an off-the-shelf inpainting model to render new objects into existing scenes. 
We find that this approach outperforms recent image editing model for object insertion,  highlighting the broader impacts of location models, which can potentially benefit downstream tasks, such as automated content creation, synthetic dataset construction, and planning for robotics. 

In summary, our main contributions are:
\begin{itemize}
    \item We propose a generative transformer model that effectively identifies plausible object placement locations, naturally handling data sparsity and multimodal outcomes. 
    \item We leverage direct preference optimization as a method to incorporate negative labels, improving bounding-box prediction accuracy in when negative labels are available.
    \item We demonstrate through extensive experiments that our method significantly surpasses existing techniques in terms of accuracy and realism, even in challenging object insertion scenarios.
\end{itemize}

%% file: floats/figures/intro_example.tex
\begin{figure}[t]
\centering
\includegraphics[width=0.9\linewidth]{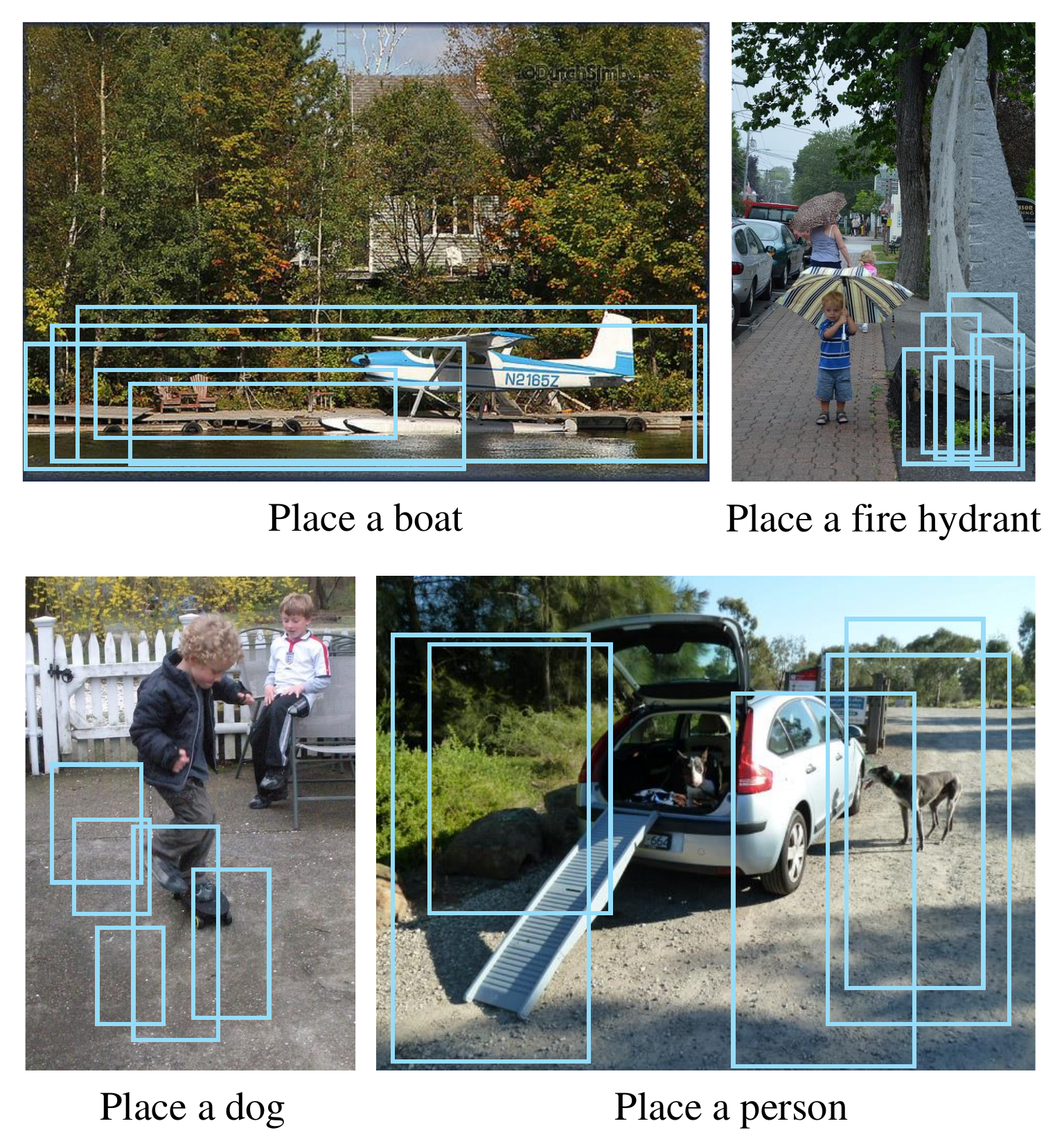}
\caption[Intro example]{A location model identifies plausible ``empty spaces'' for a hypothetical object, given the name of the object. Images taken from the OPA dataset\footnotemark.
}
\label{fig:intro-example}
\end{figure}

%% file: sections/2_background.tex
\section{Related Work}

\input{floats/figures/method}

\paragraph{Object Placement.}
Traditionally, object placement has relied on copy-pasting an object segment by simply determining its location and scale~\citep{placenet,topnet,graconet,tripathi2019learning,fopa}.
However, this approach is limited because it assumes the object appearance and shape are already defined, which is often not the case when integrating new objects into diverse and dynamic environments.

To avoid relying on object segments, some approaches predict locations from class labels by querying a classifier~\citep{contextcnn} on random bounding boxes, categorizing them as either plausible or implausible locations for the given class. 
To enable training a discriminative model (\eg, classifiers or object detectors), unlabeled locations are typically treated as negatives or implausible locations. 
Such an assumption is not always valid in object placement, as the lack of annotations for specific locations does not necessarily indicate they are implausible. 
Therefore, penalizing them may result in an inaccurate location model.
To overcome this limitation, we instead propose a generative approach that requires only samples from the target distribution, and does not make any assumption about unlabeled locations and only requires samples of the target distribution (\ie, positive locations). 

\paragraph{Layout Generation.} 
Another related category of work is focused on scene layout generation, usually via a generative model of bounding box locations~\citep{jyothi2019layoutvae, gupta2021layouttransformer, chai2023layoutdm, inoue2023layoutdm} or segmentation maps~\citep{lee2018context}.
Such layouts are often used as a condition for image generation models such as GLIGEN~\citep{gligen} to generate complex scenes~\citep{feng2024layoutgpt, phung2024grounded, cho2024visual, lian2023llm, ganillm, feng2024ranni}.
Unlike these full-layout generation approaches, our goal is to insert an object into an existing scene.
Having access to the background image fundamentally changes the nature of the task. 
On one hand, the background image provides valuable context for realistic object placement, but on the other hand, it imposes strong constraints on where objects can be placed.
We therefore design a location model that integrates these contextual cues and avoids placement in unrealistic locations.

\paragraph{Instruction-based Image Editing.}
Some works choose to solve reasoning about object interactions implicitly, in context of another task, for example image editing.
InstructPix2Pix~\citep{instructpix2pix} introduced an approach that fine-tunes Stable Diffusion~\citep{ldm} to interpret text instructions, trained with a dataset of paired images before and after specific edits. 
Subsequently, other methods~\citep{hive, hqedit, magicbrush} have released similar datasets for instruction-tuning, capable of changing the style and content of a given image.

Only recently has there been an emphasis on adding objects using text instructions, supported by datasets that provide paired images where specific objects have been artificially removed by inpainting~\citep{pbi, diffree} or using training-free techniques~\citep{addit}. 
One downside of training on such a dataset is that the data contains inpainting artifacts, which can cause models trained on them to replace existing objects or alter backgrounds. 
Additionally, since these models are typically trained to regenerate the entire image, they often introduce unintended changes to the scene. 
In contrast, a reliable location model enables us to decouple the object insertion process by first determining placement using the location model and render the object in that location using a inpainting model.
This factorization allows for more control and precision in object insertion. 

%% file: floats/figures/method.tex
\begin{figure*}
    \centering
    \includegraphics[width=\linewidth]{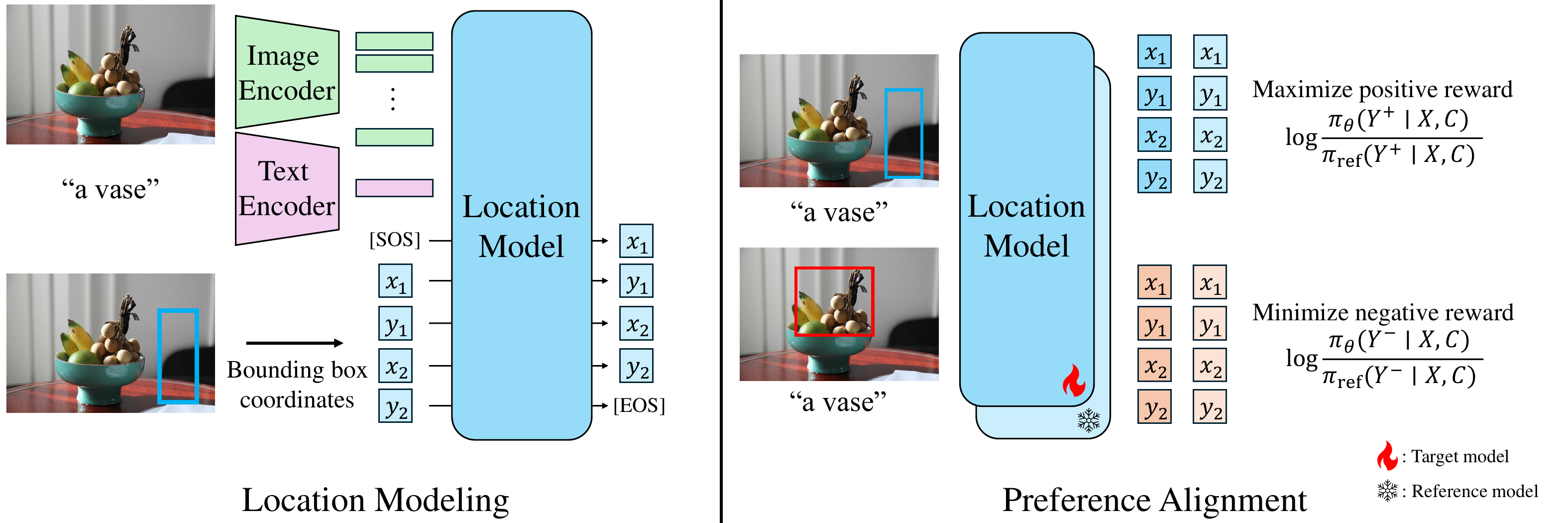}
    \caption{Training scheme of our generative location model during pretraining (left) and direct preference optimization (right). }
    \label{fig:method-main}
\end{figure*}

%% file: sections/3_method.tex
\section{Method}

\subsection{Generative Location Modeling}
Given the distribution of image $X$, plausible object locations $Y$, and classes $C$, we frame the location modeling problem using a \emph{generative model}, which estimates the conditional probability of the \emph{locations} as
\begin{equation}
\label{eq:independence}
    P(Y\mid X, C) = \prod_{Y_i\in Y} P(Y_i\mid X, C),
\end{equation}
where $Y_i$ are different locations for an object of class $C$ to be placed within the image $X$. 
Note that unlike discriminative models $P(C\mid X, Y)$, which require labels for both positive and negative locations to classify a given location, a generative model only requires samples of positive locations. 

To model this generative process, we train an autoregressive model~\citep{transformers} that sequentially predicts the bounding box coordinates of plausible locations. 
Specifically, each location $Y_i$ is represented as a bounding box with four components $\left[b_1^i,\,b_2^i,\,b_3^i,\,b_4^i\right] = \left[x_1,\,y_1,\,x_2,\,y_2\right]$, representing the coordinates of the top-left and bottom-right corners.
Thus, given a dataset $\mathcal{D}$ which provides pairs of images $X$ and plausible locations $Y$ for object category $C$, we train the model using a negative log-likelihood objective:
\begin{equation}
\label{eq:main_objective}
    \Ls_{\text{train}} = -\E_{(X,Y,C)\sim\mathcal{D}}\,\sum_{Y_i\in Y} \left[\,\sum_{k=1}^{4}\,\log P(b^i_k\mid b^i_{<k}, X, C)\,\right].
\end{equation}
where each bounding box coordinate $b^i_k$ is sequentially predicted, conditioned on previous coordinates $b^i_{<k}$, the image $X$, and the object class $C$. 

It is important to note that at training time, we model multiple bounding boxes independently (\Eqref{eq:independence}) by predicting a \textit{single bounding box} (\ie, four coordinates) for a given input. 
In other words, we sample a single location $Y_i$ from $Y$ during training. 
This choice allows us to avoid issues related to the ordering of multiple plausible bounding boxes and arbitrary sequence lengths due to the sparsity of the annotations. 
During inference, we are still able to produce multiple locations by independently sampling multiple times. 

The model architecture and the training procedure are illustrated in \cref{fig:method-main}. 
We tokenize images and the class embeddings using a pre-trained Vision Transformer (ViT)~\citep{vit} and a CLIP encoder~\citep{clip}. 
We encode bounding box coordinates by quantizing them to a grid with equally spaced bins of 1 pixel wide, (\ie, $512$ location tokens for $512\times512$ images). 
The image tokens and the target class token are prepended to the sequence, and our location model is trained to predict the probability of each coordinate in an autoregressive manner.
\subsection{Leveraging Negative Labels via DPO}
While the training objective in \Eqref{eq:main_objective} allows training the location model on sparse positive annotations, training solely on positive feedback can lead to predictions in implausible locations.
Incorporating negative annotations, whenever available, into the training objective can be beneficial for refining the model, encouraging it to assign lower likelihoods to undesirable locations and thereby improving overall accuracy.

Our generative formulation allows us to use any negative labels in the dataset as well, so that the model can learn to avoid predicting bounding boxes for implausible locations. 
Specifically, we treat the positive and negative labels as a preference dataset, where positive locations are implicitly preferred over negative ones, even though annotators were not explicitly asked to rank them. 
Using this preference structure, we fine-tune the model with direct preference optimization (DPO)~\citep{dpo}, penalizing high logits assigned to negative labels. 
We repeat the training objective below and refer the reader to \citet{dpo} Eq. 1-6 for the full derivation.

Given a target location model $\pi_\theta$ (\ie the model currently being finetuned by DPO), a reference location model $\pi_{\text{ref}}$ (\ie a frozen model trained by \Eqref{eq:main_objective} only), and a preference dataset $\mathcal{D}_\text{DPO}$ where locations $Y^+$ are preferred over $Y^-$, we can derive the likelihood of $Y^+$ being preferred over $Y^-$, based on how well the target location model  $\pi_{\theta}$ predicts each location relative to $\pi_{\text{ref}}$, following the Bradley-Terry model~\citep{bradelyterry}:
{\small
\begin{multline}
    P(Y^+\succ Y^-\mid X, C) = \Biggl[1 + \exp \Biggl(  \\
          \beta \left(\log\frac{\pi_\theta(Y^-\mid X, C)}{\pi_{\text{ref}}(Y^-\mid X, C)}
        - \log\frac{\pi_\theta(Y^+\mid X, C)}{\pi_{\text{ref}}(Y^+\mid X, C)}\right)
    \Biggr) \Biggr] ^ {-1}.
\label{eq:DPO_preference}
\end{multline}
}
Here, $\pi_\theta$ and $\pi_\text{ref}$ output the logits for the target model and the reference model, respectively, and $\beta$ is a hyperparameter. 
We can maximize the preference of $Y^+$ locations by initializing the target and reference models from a pre-trained location model, and then optimizing the target model using a negative log-likelihood objective:
{\small
\begin{multline}
    \Ls_\text{DPO} = -\E_{(Y^+,Y^-,X,C)\sim \mathcal{D}_\text{DPO}} \Biggl[ \log \sigma \Biggl(
    \\
    \beta\log\frac{\pi_\theta(Y^+\mid X, C)}{\pi_{\text{ref}}(Y^+\mid X, C)} 
    -\,\beta\log\frac{\pi_\theta(Y^-\mid X, C)}{\pi_{\text{ref}}(Y^-\mid X, C)}
    \Biggr) \Biggr].
    \label{eq:DPO_loss}
\end{multline}
}
In this way, we are able to leverage any available negative labels in the object placement dataset, and thereby improve the accuracy of the location model.

%% file: sections/4_experiments.tex
\section{Experiments}

\input{floats/figures/datasets}

\subsection{Datasets and architecture}
\label{sec:exp-details}
\paragraph{PIPE Dataset.}
The PIPE dataset~\citep{pbi} was created by removing objects from object detection datasets~\citep{coco,openimages,lvis} by inpainting.
This process results in pairs of images, one including the object and the other without it. 
To train our location model we need positive bounding box locations, that we derive for the missing object by thresholding the pixel-wise difference between the two images. 
An example is reported in \Cref{fig:datasets} (top).
While the dataset offers a large number of 888,000 samples, many images contain inpainting artifacts, potentially introducing noise in the bounding box extraction process.
\footnotetext{PIPE images are under cc-by-4.0 license: \url{https://huggingface.co/datasets/choosealicense/licenses/blob/main/markdown/cc-by-4.0.md}}

\paragraph{OPA Dataset.}
The OPA dataset~\citep{opa} was created by asking human annotators to judge the plausibility of object placement locations for a subset of COCO images~\citep{coco}.
This dataset includes on average 41.5 annotations per image, and can be used to encourage diversity in location model predictions (see \Cref{fig:datasets} (bottom) for an example). 
The train set includes 1022 images, and the test set include 130 images. 
As OPA provides negative labels, we also use OPA as the preference dataset for DPO training. 

\paragraph{Implementation details.}
We use a small GPT-2~\citep{gpt2} architecture as our autoregressive location model. 
For tokenizing the image and the object class, we use a ViT~\citep{vit} model pre-trained on ImageNet-21K~\citep{imagenet21k} as our image encoder and the CLIP text encoder~\citep{clip} as our object class encoder. 
We pre-train on PIPE for 30K iterations and fin-etune on OPA for 3K iterations. 
For batch size 64, the model can be trained on a single Nvidia V100 GPU.
We quantize each box coordinate into one of 512 bins (\ie, one bin per pixel).
For DPO training, we train using the OPA dataset for 4K iterations. 
Please refer to \Secref{sec:appx-archi} of the Appendix for further details. 

We evaluate the performance of our generative location model in an object insertion pipeline in \Secref{sec:object_insertion}, where we rely on PowerPaint~\citep{powerpaint} as the inpainting approach.
Specifically, we utilize the V2 version\footnote{\vspace{-0.5cm}https://github.com/open-mmlab/PowerPaint}, built on top of BrushNet~\citep{brushnet}.
We remark that our pipeline has the flexibility to incorporate any inpainting method for performing localized edits (see later in~\Secref{sec:ablation}).

\input{floats/figures/evaluation}

\subsection{Location Modeling}
\label{sec:exp-location}
\paragraph{Baselines. }
We compare our generative location model against discriminative approaches that classify locations as plausible or implausible. 
First, we train ContextCNN~\citep{contextcnn}, a classifier designed to assess masked regions of an image, determining their suitability for object placement. 
To be able to use it with the OPA class space, we retrain the model on COCO (whose classes comprehend all the ones in OPA).
Furthermore, we train a Faster-RCNN object detector~\citep{fasterrcnn} using positive OPA annotations to serve as a high-performing discriminative model. 

Additionally, we compare to two object placement baselines~\citep{graconet,fopa} that perform placement of specific object segments rather than generic class labels.
We use the official implementations relying on foreground segments available within the OPA dataset.
Since these methods are provided with the ground-truth object image and its aspect ratio, they are not a perfect baseline for our task. 
Nonetheless, we include their results for completeness as they also operate on the OPA dataset.

\paragraph{Evaluation Metrics.}
The OPA test set provides plausible (positive) and implausible (negative) locations for objects given an image. 
However, due to the sparse nature of these annotations, it is impossible to sample locations until every ground-truth bounding box is matched with a prediction. 
Therefore, we evaluate location models based on a ``hit rate'' metric, which compares the rate of each predicted box being a plausible or implausible location.
Specifically, we measure the True Positive Rate (TPR) and False Positive Rate (FPR) for a given set of predicted locations. 
Given $K$ predicted locations, we match them to the ground-truth labels using the Hungarian algorithm~\citep{hungarian}, where the cost function is the inverse of the intersection over union (IoU). 

True positive predictions are defined as predictions assigned to positive labels with an IoU above 0.7, while false positive predictions are those assigned to negative labels under the same IoU threshold. 
Any positive or negative ground-truth locations that are not matched are counted as false negatives and true negatives. 
Predictions that do not correspond to any labeled locations are ignored, as their true labels cannot be determined.
We refer the reader to \Cref{fig:location_modeling_results} for an example of such assignments.
TPR and FPR are then computed using standard definitions:
\begin{equation}
   \text{TPR} = \frac{\text{TP}}{\text{TP} + \text{FN}},\quad \text{FPR} = \frac{\text{FP}}{\text{FP} + \text{TN}}.
\end{equation}
Intuitively, TPR represents the rate of a predicted location being a correct (positive) location, and FPR represents the rate of a predicted location being an incorrect (negative) location.

\input{floats/figures/opa_tprfpr}
\paragraph{Results.}
We plot the TPR and FPR for different number of sampled locations $K=10, 20, \dots, 100$ in \Cref{fig:location_modeling_results}. 
Our generative location model consistently achieves a higher TPR at the same FPR, appearing in the top-left region of the plot. 
In contrast, discriminative baselines, ContextCNN~\citep{contextcnn} and FasterRCNN~\citep{fasterrcnn} fail to reach a high TPR..
This is likely due to the penalization of unlabeled positive locations during training, which limits their ability to generalize effectively. 
Our generative approach, on the other hand, avoids assumptions about unlabeled locations, enabling it to identify plausible placements with greater accuracy. 
This result highlights the effectiveness of generative modeling, particularly in scenarios where annotation sparsity hinders the training of discriminative models.

We additionally compare two object placement methods, FOPA~\citep{fopa} and GracoNet~\citep{graconet}. 
Unlike our setting, these approaches do not predict bounding boxes given object classes (which have four degrees of freedom: the coordinates of the bounding box), but rather predict the location and scale of an object (with three degrees of freedom) given an image of the object.
Despite having access to the ground-truth object segments and aspect ratios, both FOPA and GracoNet fail to outperform our approach, further demonstrating the advantages of our generative model in producing more accurate and plausible placements.

\input{floats/figures/quantitative_opa}

\subsection{Object Insertion}
\label{sec:object_insertion}

\paragraph{Baselines. }
We evaluate our location model in an object insertion pipeline, where we first identify plausible insertion locations using our generative location model and subsequently render objects at these locations using an off-the-shelf inpainting model, PowerPaint~\citep{powerpaint}.
Our method is evaluated against several strong baselines, including general instruction-based image editing models such as InstructPix2Pix~\citep{instructpix2pix}, HIVE~\citep{hive}, and MagicBrush~\citep{magicbrush}, as well as specialized object-insertion models like Paint-by-Inpaint~\citep{pbi} and Diffree~\citep{diffree}. 
To further isolate and demonstrate the advantages of our generative location model, we additionally experiment with Faster-RCNN~\citep{fasterrcnn}, a discriminative location model, integrated into the same insertion pipeline.

These comparisons illustrate that explicitly modeling plausible object locations significantly enhances insertion realism and accuracy, providing a clear advantage over general-purpose editing approaches and other less accurate location models. 

\paragraph{Evaluation Metrics.}
Similar to existing image editing benchmarks~\citep{pbi}, we evaluate the success of object insertion by considering both how well the original scene is preserved and how accurately the target object is inserted. 
Using an object detector~\citep{detr} we separate the background, which ideally should remain untouched, and the foreground, where the newly inserted object appears. 

To measure background preservation, we compute the Structural Similarity Index Measure (SSIM)~\citep{ssim} on the background region. 
To assess the accuracy of the object, we measure the CLIP similarity~\citep{clip} between the cropped object and the text ``\textit{an \{object class\}}''.
If no new object is detected, we assign a CLIP score of 0 to reflect the failure to properly insert the object. 

For the PIPE dataset, we also evaluate the diversity of edits by calculating the average LPIPS distance~\citep{lpips} across 10 different edits per background-object pair. 
High LPIPS indicates that a model can generate diverse results from the same instruction, highlighting approaches that are limited to producing a single edit.

\paragraph{Results on OPA.}
We compare instruction-finetuned image editing models, location models paired with strong inpainting models, and our approach in \Cref{fig:exp-img-curve}.
Our approach substantially outperforms all baselines by leveraging a dedicated location model for inpainting, which effectively reduces background distortions while maintaining high-quality object generation. 

By design, localized inpainting is an effective strategy for preserving the background, but it is notable that even a random location model outperforms the best instruction-finetuned method on background SSIM. 
Since the key difference between our approach and other location modeling baselines lies in the plausibility of the predicted locations, the figure suggests that an accurate placement directly impacts the quality of inpainting results. 

The instruction-tuned models often face a trade-off. 
They either preserve the background well but fail to generate the object convincingly, or they successfully generate the object but significantly alter the surrounding scene. 
This inconsistency may stem from the fact that these models address both the placement and generation tasks simultaneously, leading to suboptimal object locations. 
In contrast, using a dedicated location model allows one model to focus on spatial reasoning, while the inpainting method concentrates on rendering realistic objects. 
\input{floats/tables/pipe_quantitative}
\paragraph{Results on PIPE. }
We further evaluate our approach on the PIPE test set~\citep{pbi}, which is a dedicated dataset for evaluating object insertion methods. 
Using the same metrics reported in Table~\ref{tab:pipe_quantitative}, our results are consistent with those observed in the OPA test set. 
Notably, Paint-by-Inpaint~\citep{pbi} lacks diversity, frequently generating identical outputs, as seen in extremely low LPIPS scores. 
This is likely due to training on datasets created by object removal through inpainting.
Moreover, although we appreciate how InstructPix2Pix and HIVE achieve very diverse edits, we notice that this result is typically achieved by generating entirely new images, rather than editing the input scene, as also testified by their low background SSIM scores.
\paragraph{User Study. }
To evaluate the quality of object placement from a human perceptual perspective, we conducted a user study comparing our method against four competitive baselines.
We presented 46 participants with pairs of edited images from the OPA dataset, asking each participant to select the image with the more realistic and coherent object insertion. 
Each participant evaluated 40 image pairs. 
Further details can be found in Appendix \Secref{sec:appx-userstudy}.

The results of the user study, summarized in \Cref{fig:exp-userstudy}, clearly demonstrate that our approach is consistently preferred over the baselines.
Notably, our generative location model significantly outperforms the strongest discriminative baseline, FasterRCNN~\citep{fasterrcnn}, with twice as many participants preferring edits guided by our model's predicted locations. 
These findings underscore the critical impact of accurate location modeling, highlighting that improvements at the placement stage propagate substantially into downstream applications reliant on realistic object locations.

\input{floats/figures/user_study}
\input{floats/figures/qualitative_opa}

\paragraph{Ablation study.}
\label{sec:ablation}
Training a generative location model exclusively on positive locations already demonstrates strong performance, as shown in \Cref{fig:location_modeling_results}, Ours w/o DPO.
However, incorporating negative labels further enhances accuracy by explicitly guiding the model on where \textit{not} to predict object locations.
Notably, unlike existing techniques which assume non-labeled locations as negative even when negative labels are present, our approach leverages only the negative labels provided by annotators. 
This ensures more precise predictions, as it avoids the potential inaccuracies introduced by assuming non-labeled locations are negative.

%% file: floats/figures/datasets.tex
\begin{figure}[t]
    \centering
    \includegraphics[width=0.9\linewidth]{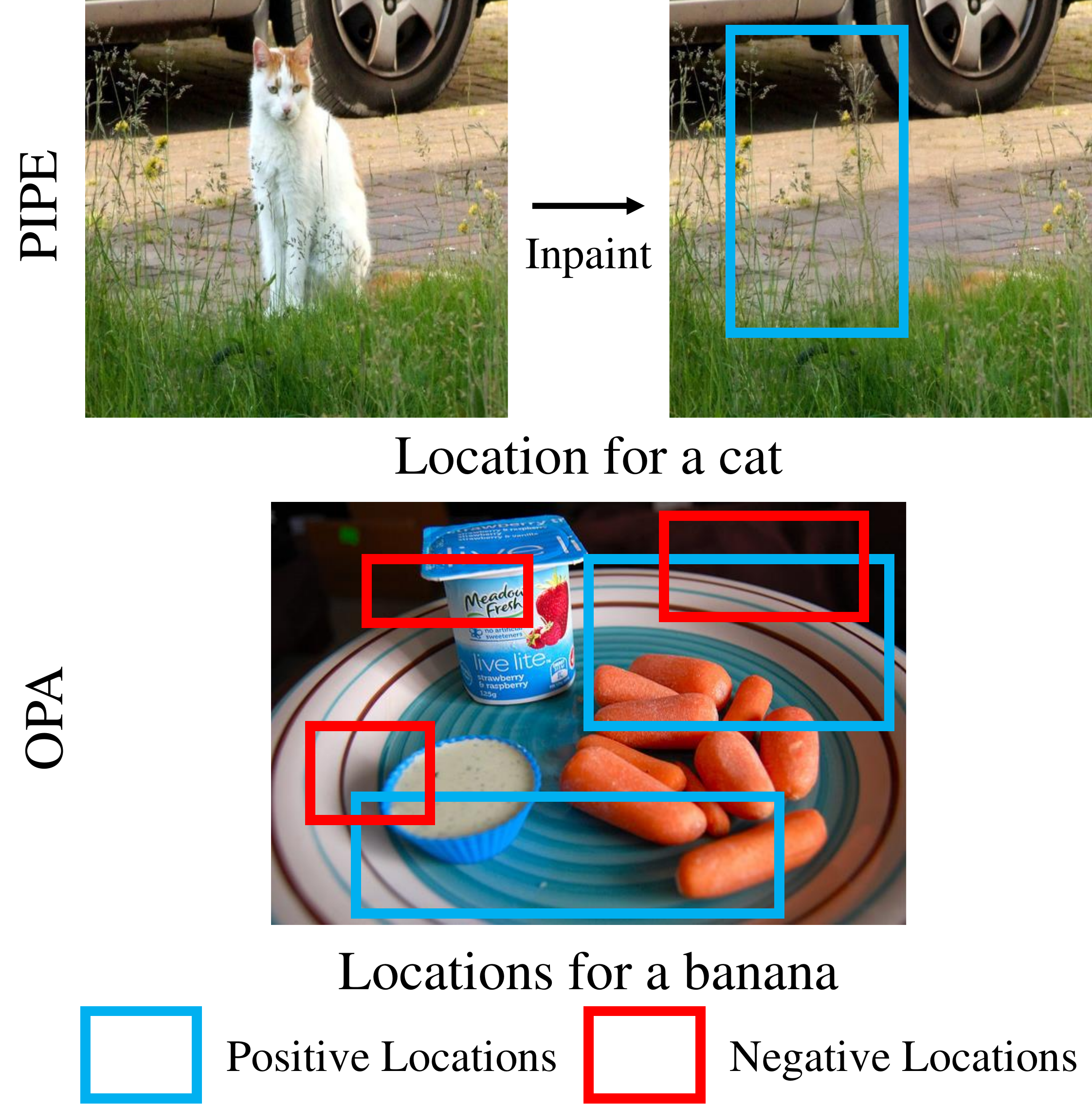}
    \caption[datasets]{Annotation format for the PIPE\footnotemark and OPA dataset. The PIPE dataset has one groundtruth location per image, whereas OPA provides multiple positive and negative locations.}
    \label{fig:datasets}
\end{figure}

%% file: floats/figures/evaluation.tex
\begin{figure}[t]
\centering
\includegraphics[width=\linewidth]{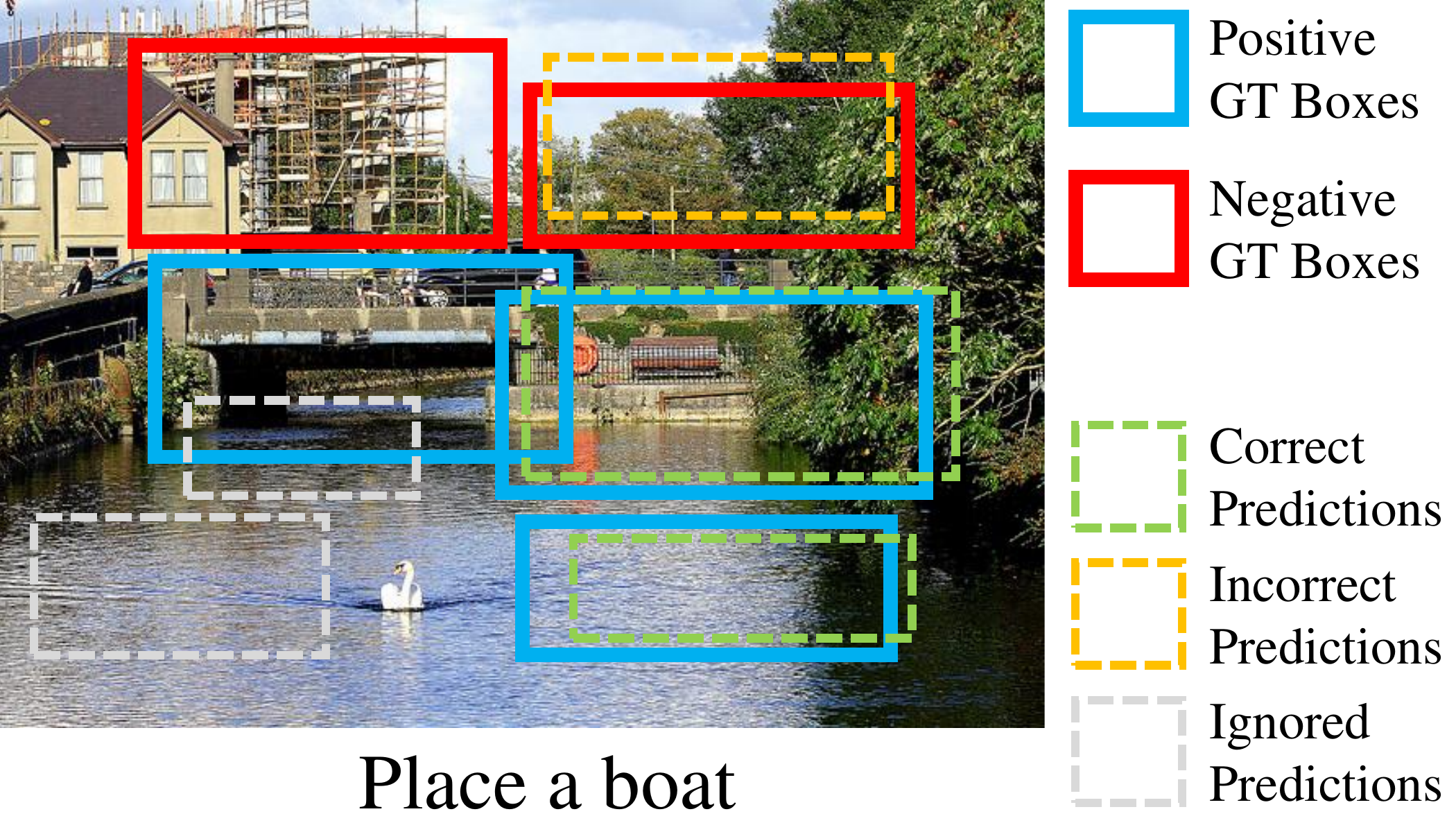}
\caption{Example evaluation scenario. Predicted boxes are counted only if a positive or negative ground-truth box meets an IoU above the threshold.}
\label{fig:evaluation}
\end{figure}

%% file: floats/figures/opa_tprfpr.tex
\begin{figure}[t]
    \centering
    \includegraphics[width=\linewidth]{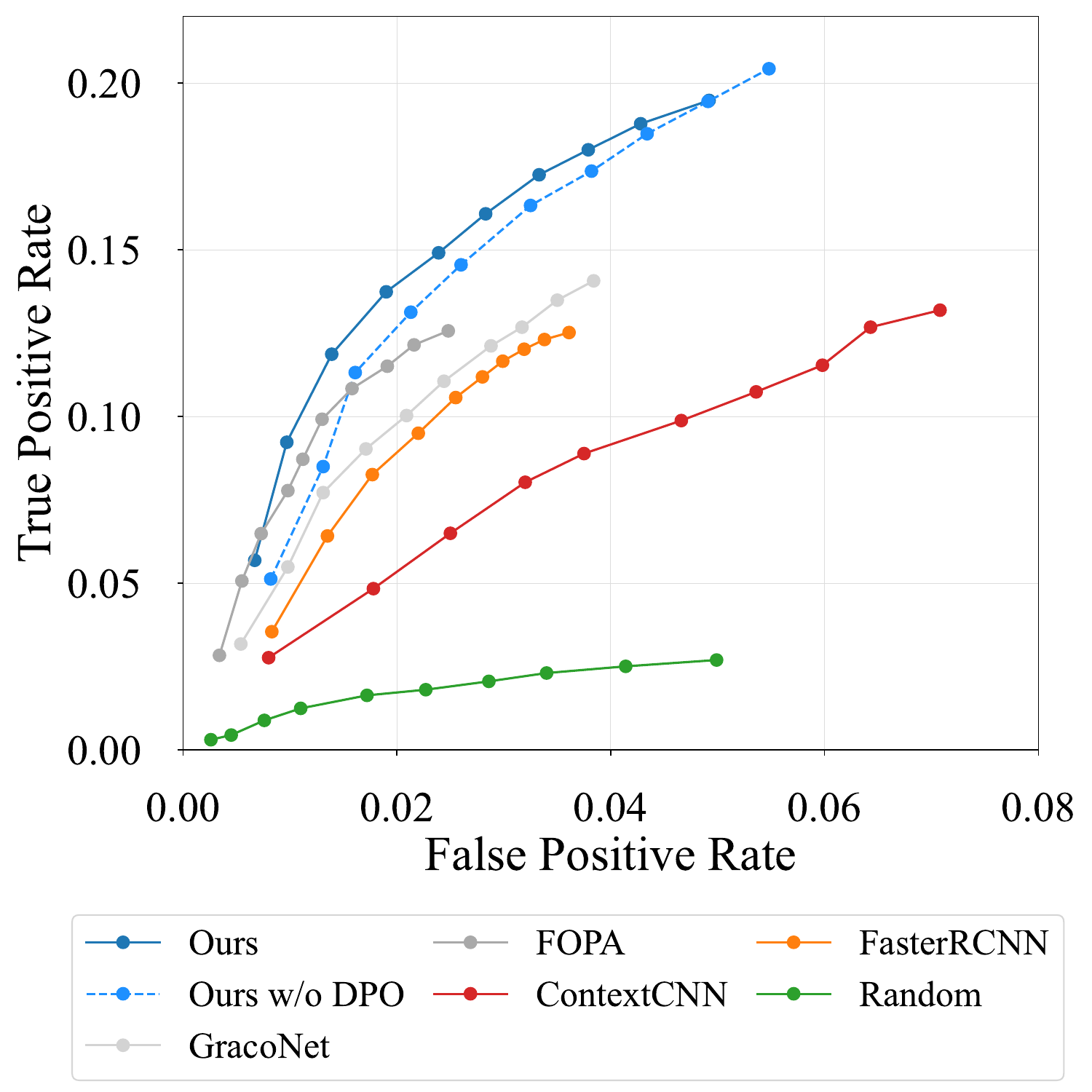}
\caption{TPR-FPR curves. Each line is constructed by sampling $\{ 10, 20, \ldots, 100 \}$ locations. Top-left is better.}
\label{fig:location_modeling_results}
\end{figure}

%% file: floats/figures/quantitative_opa.tex
\begin{figure}
   \centering
   \includegraphics[width=\linewidth]
   {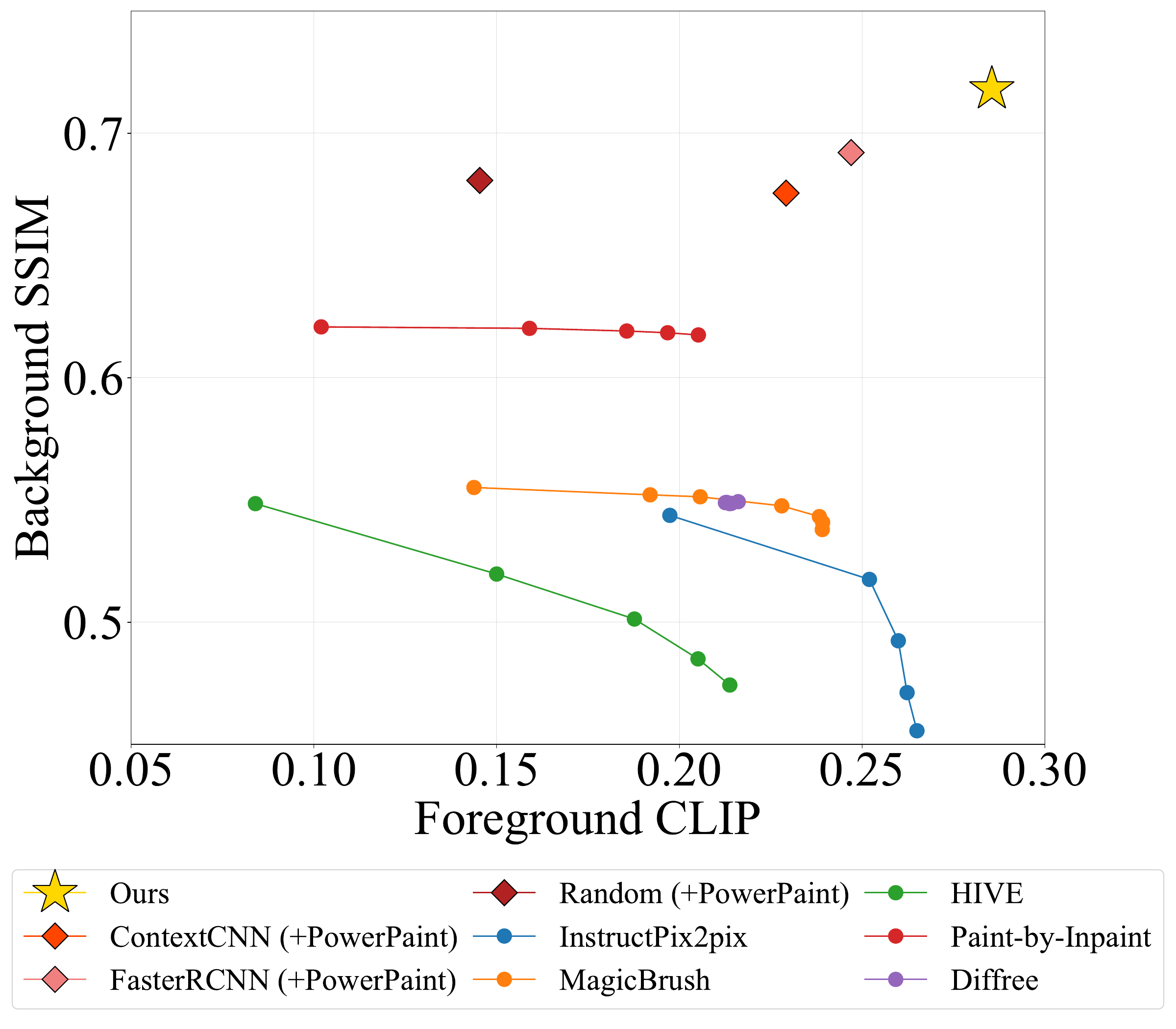}
   \caption{Quantitative evaluation on the OPA dataset, higher and to the right is best. For instruction-tuned approaches, each dot represents a different guidance scale ranging from 2 to 10. For other methods, guidance scale has a negligible effect, hence we show a single point.}
   \label{fig:exp-img-curve}
\end{figure}

%% file: floats/tables/pipe_quantitative.tex
\begin{table}[t]
\centering
\resizebox{1.0\columnwidth}{!}{
\renewcommand{\arraystretch}{1.2}
\begin{tabular}{@{}cx{1.8cm}x{1.8cm}x{1.8cm}@{}}
\toprule
                 & BG SSIM $\uparrow$& FG CLIP $\uparrow$ & LPIPS  $\uparrow$\\ \midrule
InstructPix2Pix  & 0.5679 & 0.2670 & \textbf{0.3865} \\
MagicBrush       & 0.6118 & 0.2579 & 0.0986 \\
HIVE             & 0.5635 & 0.2047 & 0.1886 \\
Diffree          & 0.6170 & 0.2517 & 0.1210 \\
Paint-by-Inpaint & 0.7281 & 0.2672 & 0.0700 \\ \midrule
Ours             & \textbf{0.8075} & \textbf{0.2774} & 0.1824 \\ \bottomrule
\end{tabular}}
\caption{Evaluation on PIPE. BG SSIM represents background similarity, FG CLIP measures faithful generation of the object, and LPIPS measure the diversity of image across multiple generations. }
\label{tab:pipe_quantitative}
\end{table}

%% file: floats/figures/user_study.tex
\begin{figure}
\centering    
\includegraphics[width=\linewidth]{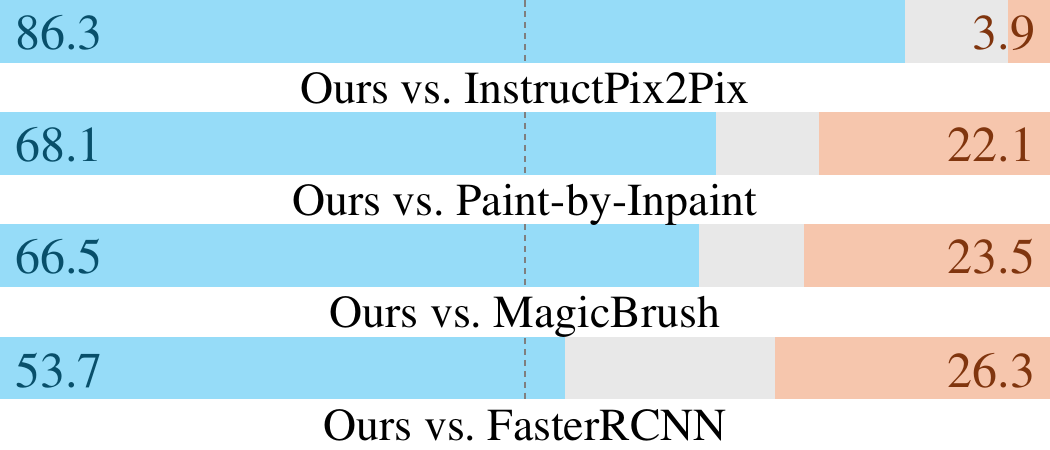}
\caption{User study on edited images (OPA). Blue: ours preferred. Red: baseline preferred. Grey: no preference.}
\label{fig:exp-userstudy}
\end{figure}

%% file: floats/figures/qualitative_opa.tex
\begin{figure*}
\centering
\includegraphics[width=0.95\linewidth]{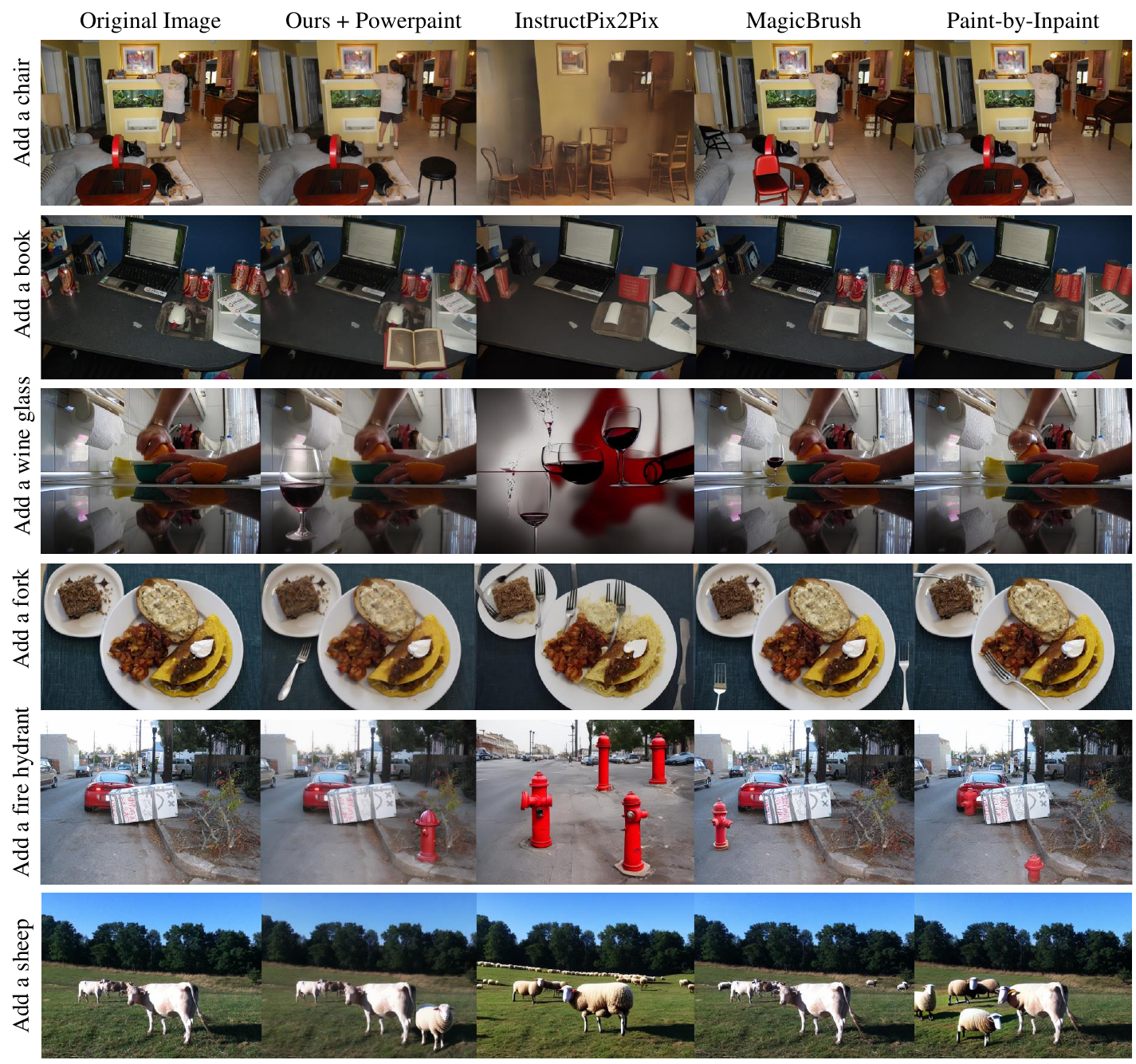}
\caption{Comparison between our method + PowerPaint, and instruction-guided image editing models on the OPA dataset. Best viewed electronically.}
\vspace{-0.3cm}
\label{fig:exp-quali-opa}
\end{figure*}

%% file: sections/5_discussion.tex
\section{Discussion}
\paragraph{Inference Cost.}
Our location model introduces a minimal overhead to the object insertion process. 
To measure it, we compare the inference time of the location model relative to the time required for rendering the image (\ie, inpainting).
On an average across 100 runs, our model takes 0.03 seconds to sample a single location on a Nvidia Tesla V100 GPU, a minor addition compared to the 7.10 seconds needed to render an image using a 50-step diffusion reverse process for a StableDiffusion v2.1~\citep{ldm}.
Paying a small upfront cost for identifying a plausible location leads to a significant improvement in the quality of object insertion.

\paragraph{Limitations.}
While our generative formulation enables effective training of location models with sparse annotations, it is important to note that these annotations remain costly to collect. 
As a result, our approach is currently limited to datasets that provide such annotations. 
A growing trend, involves the automatic generation of such data by removing objects from scenes (\eg PIPE dataset). 
In the future, we aim to expand these datasets by incorporating multiple object removals, thus enabling a more comprehensive modeling of plausible locations across various scenarios. 

%% file: sections/6_conclusion.tex
\section{Conclusion}
In this work, we propose a generative location model designed to identify plausible locations for inserting hypothetical objects into existing scenes. 
Unlike previous discriminative approaches, our generative formulation naturally addresses the inherent sparsity and ambiguity of location annotations, enabling robust and accurate predictions even when labeled data is limited. 
Our experiments demonstrate that our model effectively captures the distribution of plausible locations.
We perform additional experiments on a downstream task,  object insertion. 
These experiments confirm that accurately predicted locations significantly enhance the realism and coherence of synthesized images, underscoring the crucial role that high-quality location modeling plays in successful object insertion.

More generally, we believe that building spatial awareness is a key factor for building reliable models interacting with the real world. 
This is true whether a model operates in the image editing setting, as in this work, or in more complex domains, such as robotics or virtual reality. 
Our work shows that a model is able to effectively learn such awareness from example positive and negative annotations, following a generative modeling of these locations.
Although we focus on locations for 2D bounding boxes conditioned on images, we hope to see similar models scale to handle other scene representations (\eg, depth or semantic layouts) and precise locations (\eg, 3D bounding boxes or object masks) in the future. 

%% file: sections/7_appendix.tex
\appendix
\section*{\Large Appendix}

\section{Dataset Statistics and Preprocessing}
\label{sec:appx-datasets}
\subsection{PIPE Dataset} 
The PIPE dataset~\citep{pbi} was created by removing objects from object detection datasets~\citep{coco, openimages, lvis} using an inpainting model, resulting in over 600 object classes for insertion and more than 888,000 training pairs of images, showing scenes before and after object removal.
To preprocess this dataset, we pair each background image with the original location of the removed object. 
The locations are identified by computing the pixel-wise difference between the before-and-after images, and extracting coordinates where the difference exceeds a certain threshold. 
Despite this thresholding, the resulting bounding boxes can be noisy, and the background images generated by the inpainting model often contain artifacts. 
Also, the PIPE dataset only provides a single positive location for an object for each background image and does not provide negative labels. 

\begin{figure*}
    \centering
    \includegraphics[width=\linewidth]{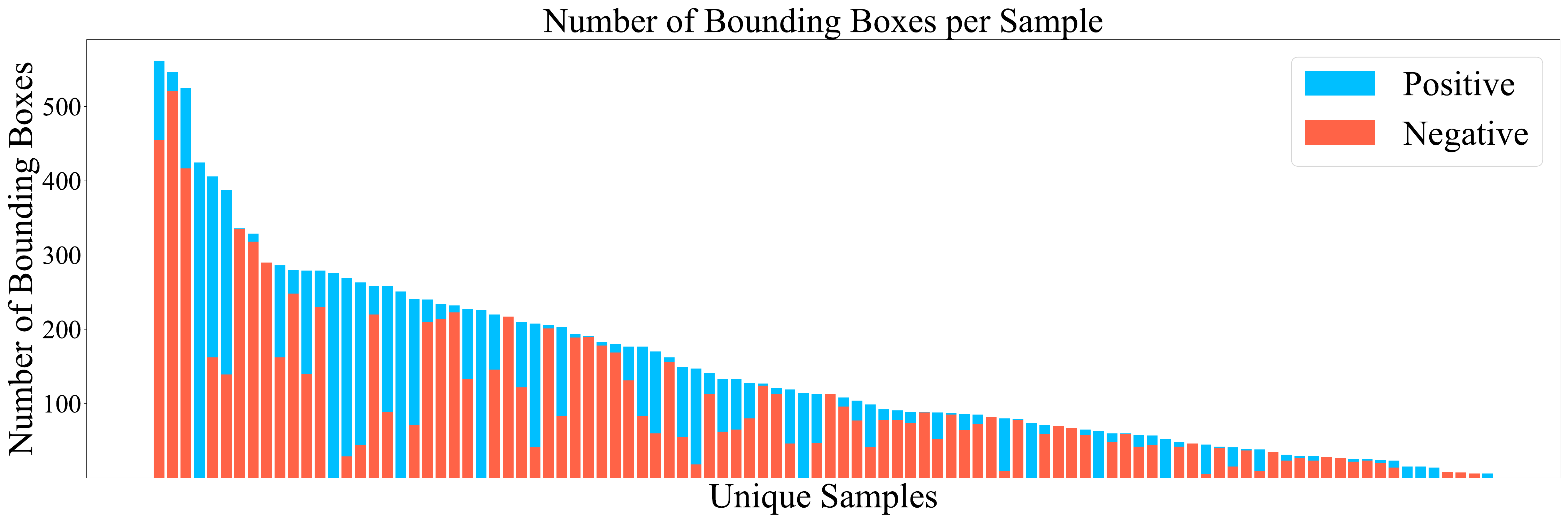}
    \vspace{-0.2cm}
    \caption{Distribution of positive and negative bounding boxes in the OPA dataset. We randomly select 100 samples (images and their annotations) from the training set for visualization.}
    \label{fig:appx-dataset-stats}
\end{figure*}

\subsection{OPA Dataset}
The OPA dataset was created by manually annotating samples from the COCO dataset~\citep{coco}, resulting in 47 object categories for object placement. 
The dataset is intended for the task of finding locations to copy-paste images of objects, and therefore includes background images, object images with transparent backgrounds, and labeled plausible/implausible locations for placing the objects.
Since our focus is on object location modeling rather than the insertion of object images, we ignore the object images and restructure the dataset to include pairs of background images and their corresponding object locations. 
This restructuring yields 1,496 training samples and 184 test samples. 
While the number of images is relatively small, each sample contains around 40 annotations, making it a richly annotated dataset. 
In total, the OPA train set contains 21,376 positive labels and 40,698 negative labels. 

Despite having an average 40 annotations for each sample, this accounts for fewer than 1\% of the typical number of anchor boxes used in object detectors, leaving most locations unlabeled. 
Furthermore, as illustrated in \cref{fig:appx-dataset-stats}, not only are the number of bounding boxes across samples highly imbalanced, but the distribution of positive and negative labels for each sample is extremely inconsistent. 
This sparsity and imbalance make it extremely challenging to train discriminative models that classify locations as plausible or implausible. 
Our generative approach bypasses this issue by modeling the distribution of plausible locations, using negative labels only for DPO.

\section{Does Location Matter?}

\begin{figure}
    \centering
    \includegraphics[width=0.95\linewidth]{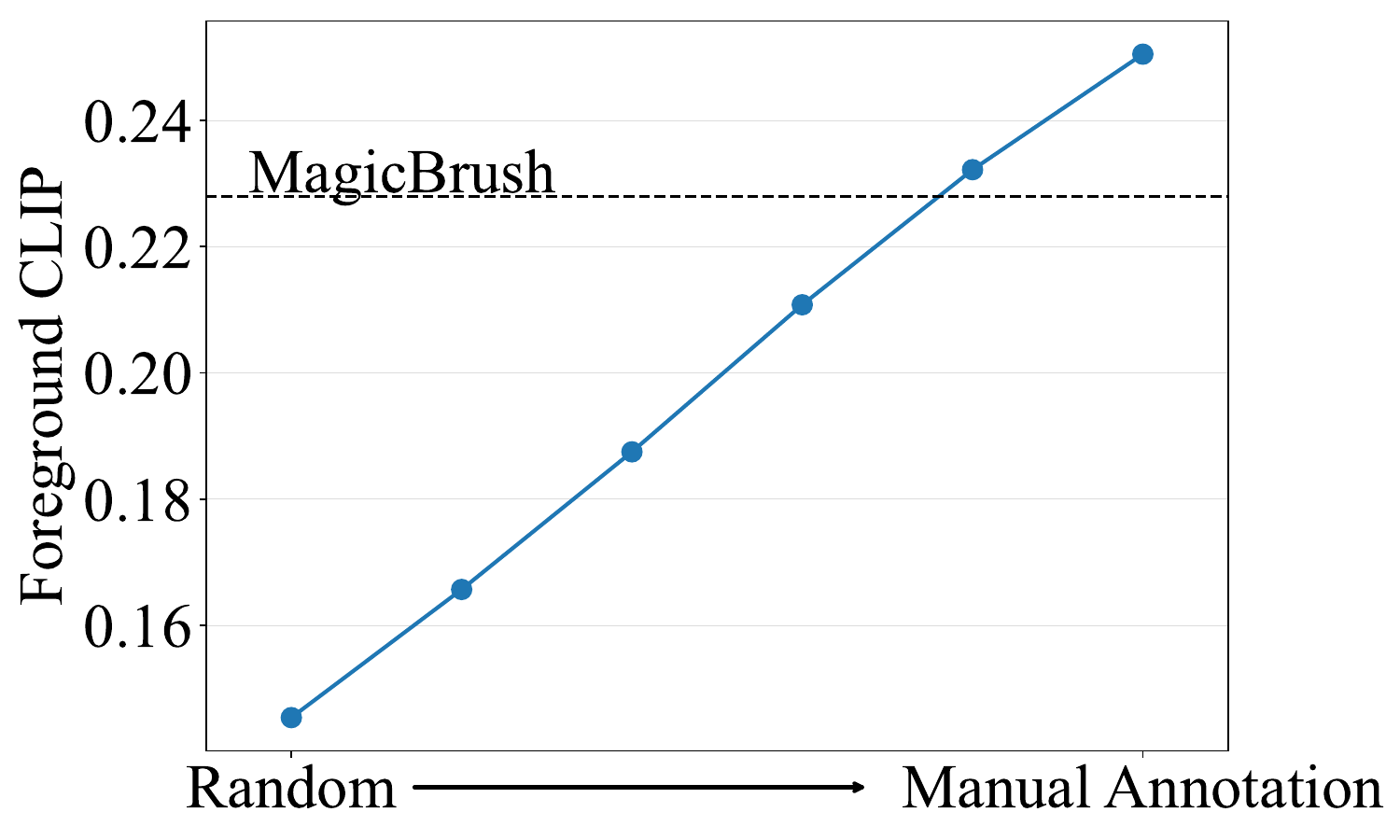}
    \caption{Inpainting quality with respect to the quality of the location. The success of inpainting approaches for object insertion is highly dependant on the quality of the location. }
    \label{fig:appx-location-clip}
\end{figure}

To demonstrate that the quality of locations directly influences the quality of generated objects, we present a proof-of-concept experiment in \Secref{fig:appx-location-clip}. 
For each image in the OPA dataset, we uniformly sample 10 random locations and then interpolate between these random locations and manually annotated ones, performing inpainting at six distinct interpolation points. 
To evaluate the success of inpainting, we measure the foreground CLIP similarity~\citep{clip} between the cropped object and the text ``\textit{an \{object class\}}''. 
For reference, we also include the performance of MagicBrush, which inserts objects without explicit location modeling. 
As the inpainting locations become more precise, the fidelity of generated objects increases, underscoring the importance of accurate locations. 
Inpainting in incorrect locations often results in failed insertions, motivating the development of a dedicated location model to provide spatial awareness for inpainting models. 

\section{VLMs for Location Modeling}
\begin{figure}
    \centering
    \includegraphics[width=\linewidth]{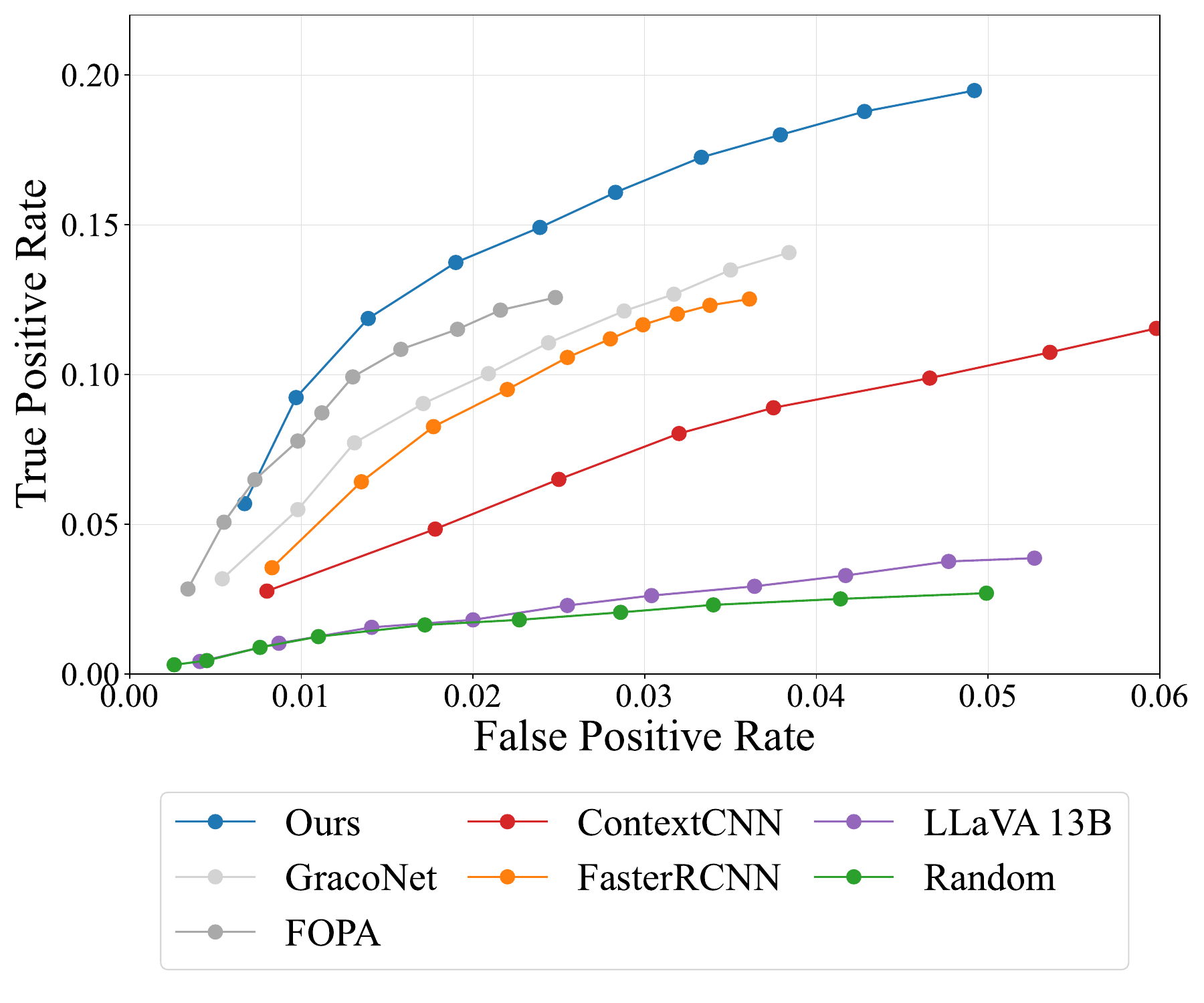}
    \vspace{-0.2cm}
    \caption{TPR-FPR curves compared with object placement approaches and LLaVA. }
    \label{fig:appx-tpr-full}
\end{figure}

Vision-language models (VLMs) can also be used to predict plausible locations for object placement based on image inputs. 
Although these models can generate bounding box coordinates as part of their responses, we find that they are largely ineffective at predicting meaningful locations, with their bounding box outputs being comparable to random guesses. 
Specifically, we use LLaVA-13B~\citep{llava} using the following prompt:
\begin{center}
\begin{lstlisting}[basicstyle=\scriptsize]
USER: <image>If a new {object_name} would appear in this 
scene, what would be the coordinates of 
the {num_samples} different plausible locations? 
Answer in JSON format 
    {
        plausible location 1\": [x1, y1, x2, y2], 
        plausible location 2\": [x1, y1, x2, y2], ...,
        plausible location {num_samples}: [x1, y1, x2, y2]
    }.
Output must only include the JSON format and no other text.
ASSISTANT: In this scene, {num_samples} most plausible 
locations of a newly inserted {object_name} are:
\end{lstlisting}
\end{center}
We also compare against object placement approaches~\citep{graconet, fopa} that predict locations based on both the background image and a tightly masked image of the object. 
Since these models are provided with the exact aspect ratio of the object, they only need to predict the location and scale of the bounding box.
As previously mentioned in \Secref{sec:appx-datasets}, the OPA dataset includes object images, allowing us to measure the TPR and FPR on the same test set. 
Despite having the advantage of having provided with ground-truth aspect ratios, these models are outperformed by our location model, which demonstrates superior performance even without access to object images.
A full comparison of these location models are plotted in \cref{fig:appx-tpr-full}. 

\section{Architecture and Implementation Details}
\label{sec:appx-archi}
\subsection{Architecture of the Location Model}
Our location model is based on a GPT-2 small architecture~\citep{gpt2}, consisting of 12 layers of Transformer blocks~\citep{transformers}. 
For image and object class encoding, we use a pre-trained ViT-B model~\citep{timm, vit} for the image encoder and a ViT-B CLIP text encoder~\citep{openclip, clip} for the object class. 
The images are converted into 196 embeddings, while the text is transformed into a single embedding, creating a sequence length of 197 when combined. 
These 197 embeddings are prepended to our location model before predicting the coordinates. 
To quantize the coordinates, we use 512 bins for both height and width, resulting in a vocabulary size of 514, including the start-of-sequence (SOS) and end-of-sequence (EOS) tokens. 

The location model, including the ViT-B backbone, comprises a total of 411 million parameters, and loading the weights in float16 precision requires just 2.05 GB of VRAM. 
Running the model with batch size 1 in float16 precision requires an additional 1.93 GB of VRAM for the activations, meaning the model can easily be run on consumer-grade hardware.
On a Nvidia Tesla V100 GPU, inference runs in 0.03 seconds.
For editing images, we first sample locations, unload the model from the GPU, and perform inpainting, ensuring no additional memory overhead beyond what the inpainting model requires.

In comparison, the PowerPaint inpainting model has three components: a VAE with 83.7 million parameters, a text encoder with 123.1 million parameters, and a UNet with 859.5 million parameters.
Although the location model is nearly half the size of the UNet in terms of parameters, it is used only once in advance (similar to the VAE and text encoder), whereas the UNet is used at every reverse step.
On the same GPU, the it takes 7.10 seconds to perform 50 reverse steps, meaning the inference overhead of the location model is 0.4\%, an acceptable cost for the improvement in generation quality.

\subsection{Training Details}
We train the model using the Adam optimizer~\citep{adam} for training the location model and Stochastic Gradient Descent (SGD) for DPO training. 
The model is trained on the PIPE dataset for 30,000 iterations with a learning rate of 1e-4, incorporating a linear warmup over the first 1,000 steps. 
Subsequently, we fine-tune the model on the OPA dataset for 3,600 steps and perform DPO for an additional 4,600 steps.
For batch size 64, the model can be trained on a single Nvidia V100 GPU.

\begin{figure}
    \centering
    \includegraphics[width=0.9\linewidth]{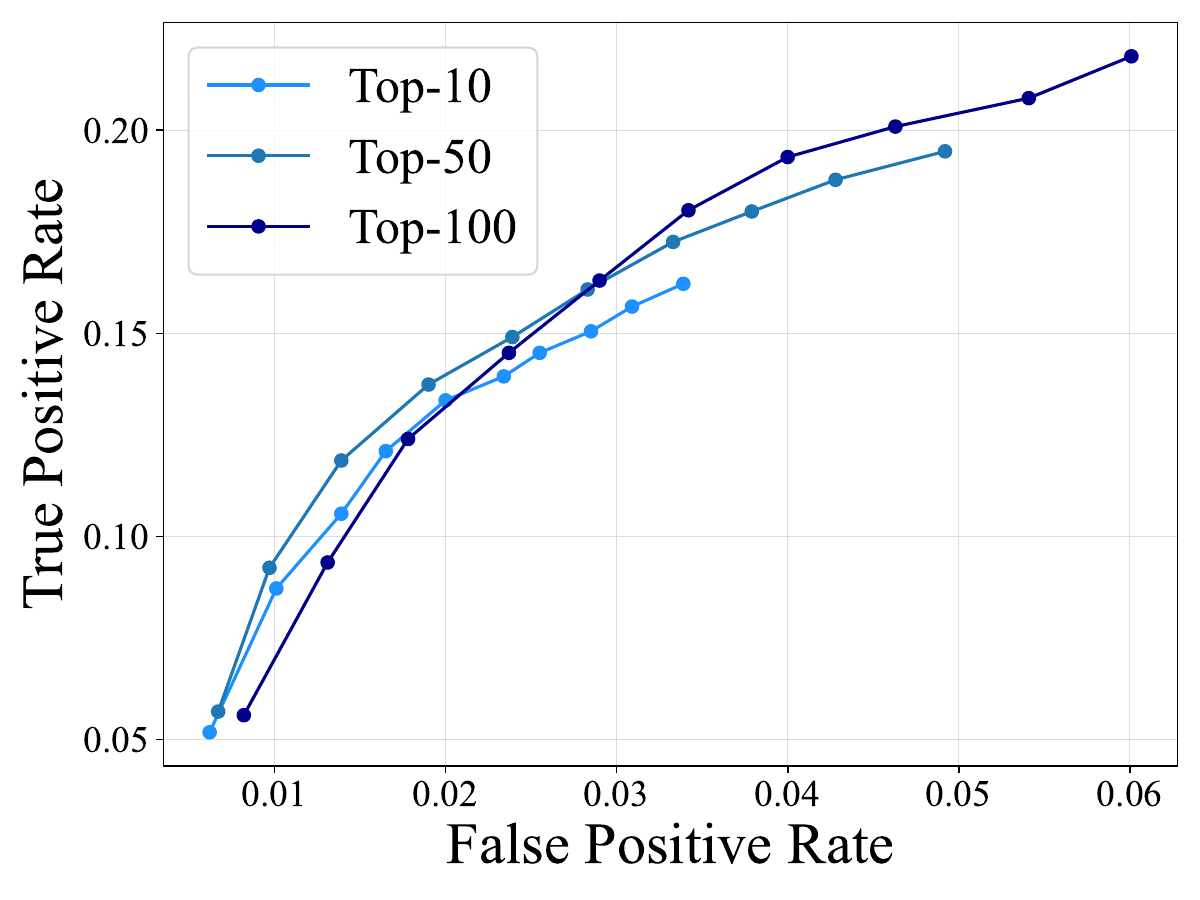}
    \caption{Location modeling performance for different top-$k$ parameters used during sampling. Higher numbers of $k$ leads to diverse predictions, often at the cost of accuracy. }
    \label{fig:appx-topk}
\end{figure}

\subsection{Sampling from the Locations}
Our autoregressive layout model can follow sampling techniques similar to those used for text generation using Large Language Models (LLMs)~\citep{gpt2}. 
We sample among the top-$k$ probabilities scaled with a temperature of $1.0$. 
As illustrated in \cref{fig:appx-topk}, we find that the higher values of $k$ promotes diversity and thus achieves a higher True Positive Rate (TPR), but also increases the False Positive Rate (FPR). 
For the main experiments, we use $k=50$ during sampling. 

\section{Diversity of the Sampled Location}

An essential aspect of location models is their ability to identify diverse plausible locations, as multiple potential placements often exist for a given object. 
To quantify this diversity, we apply Non-maximum Suppression (NMS) to the predicted locations and count the remaining bounding boxes.
Specifically, we perform a standard NMS with a threshold of 0.7 on a set of 100 predicted locations for the same object in the same scene.
We observe that 82.5 bounding boxes remain after performing NMS, which highlights the diversity of our location model.

\section{Human Evaluation}
\label{sec:appx-userstudy}
We use four baselines in our user study: three instruction-finetuned models, and the strongest location modeling baseline (\ie, Faster-RCNN trained on OPA + Powerpaint).

Participants were presented with the original image, an instruction (\textit{``Add a \{object class\}''}), and two edited images: one generated by our approach and the other by an instruction-tuned editing model or location modeling baseline. 
For each image, we randomly swap which method is shown on the left or right side.
Each participant was asked to evaluate which of the two edited images better adhered to the editing instruction and maintained the overall coherence of the scene. 
In total, we collected 1,840 responses from 46 participants, with each individual comparing $4 \times 10$ pairs of randomly selected samples.

\section{Qualitative Results}
\label{sec:appx-quali}

We provide additional image editing results from the PIPE dataset. 
In our observations, instruction-tuned models often lack diversity in their final edits when they successfully insert objects. 
When these models fail to insert objects, they either leave the image unchanged or modify too much of the scene, which are both regarded as a failure to insert objects.

\begin{figure*}
    \centering
    \includegraphics[width=0.93\linewidth]{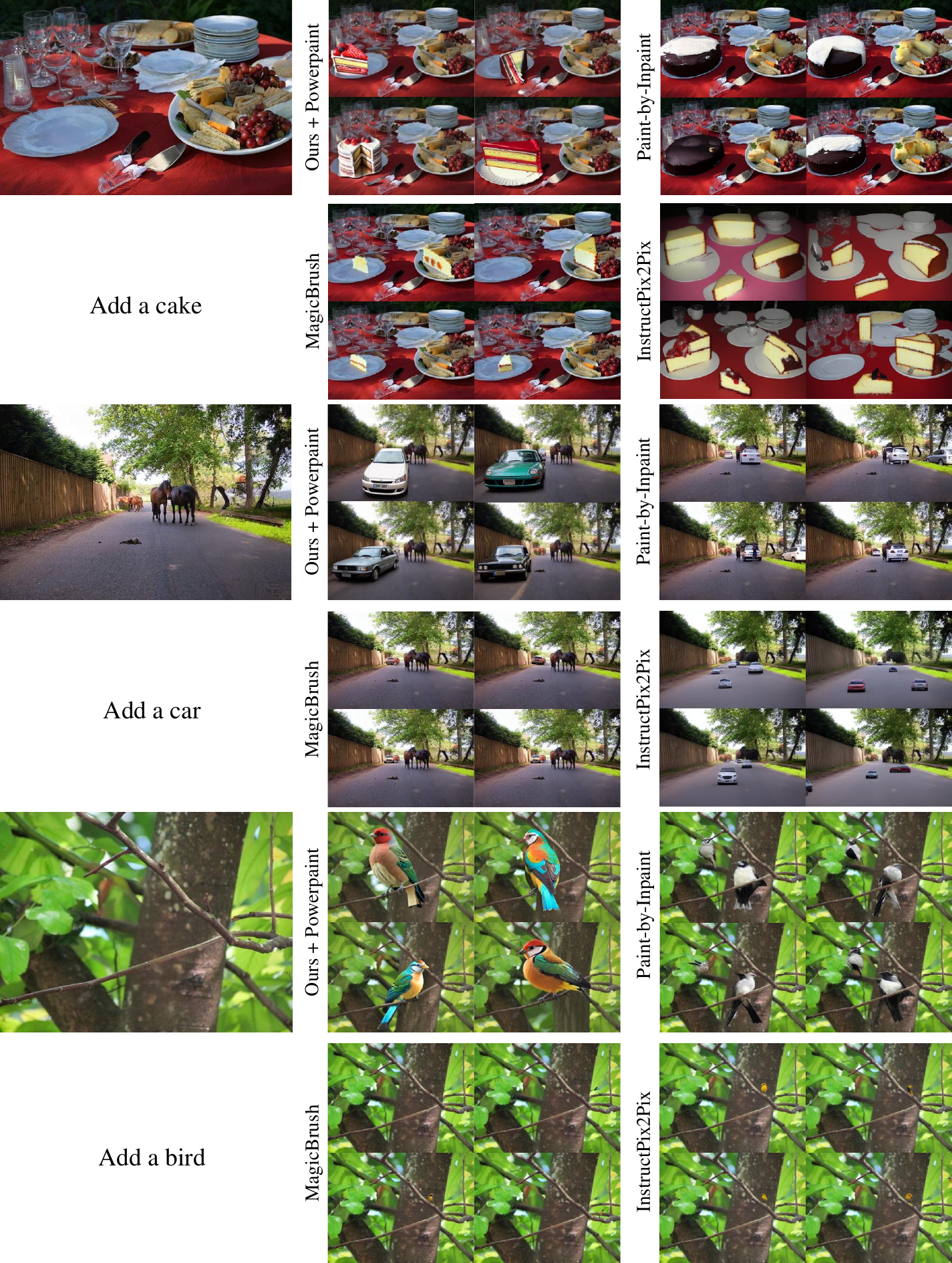}
    \caption{Additional samples of object insertion results in the PIPE test set.}
    \label{fig:appx-quali-pipe1}
\end{figure*}
\begin{figure*}
    \centering
    \includegraphics[width=0.89\linewidth]{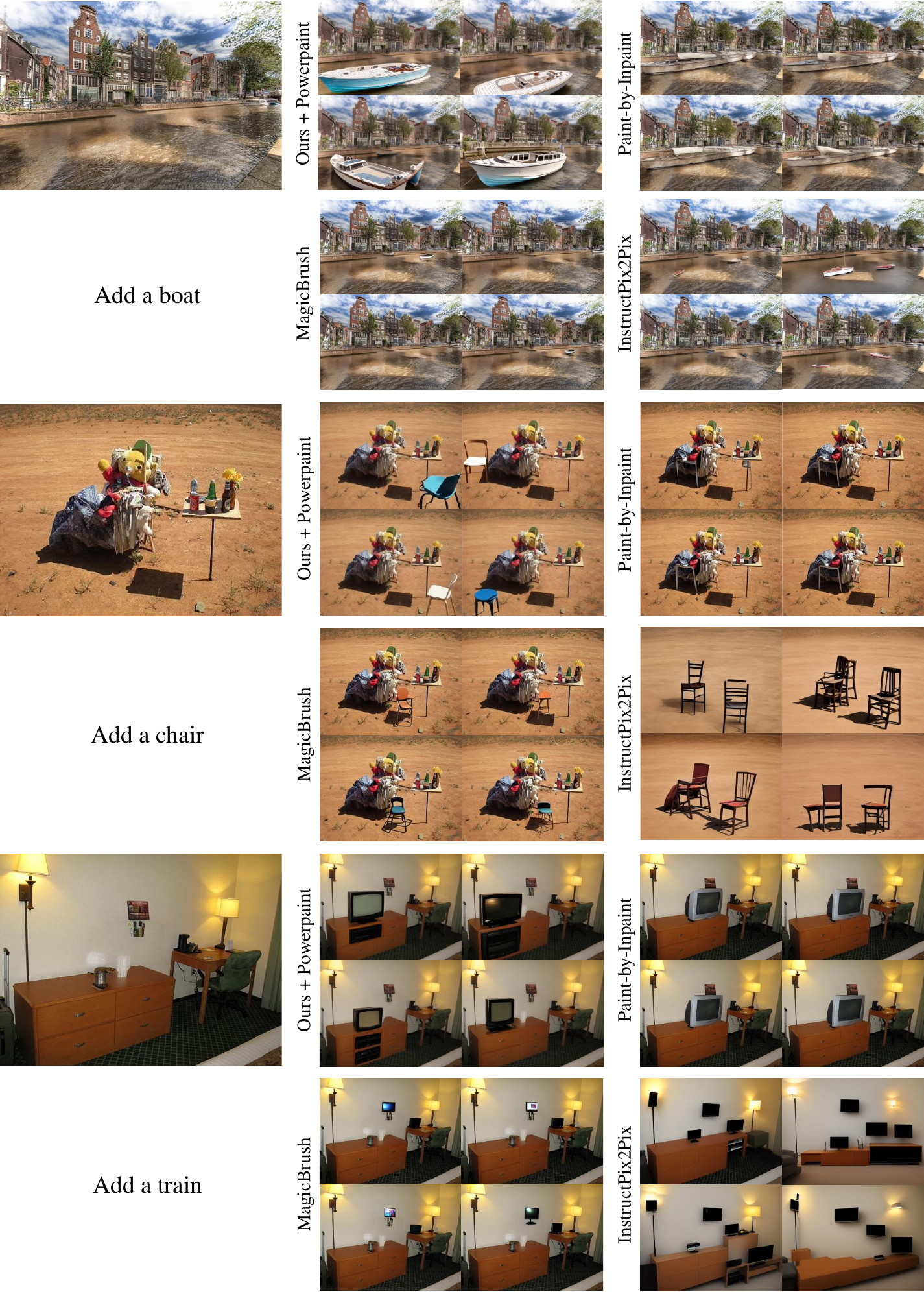}
    \caption{Additional samples of object insertion results in the PIPE test set.}
    \label{fig:appx-qauli-pipe2}
\end{figure*}
\begin{figure*}[t]
    \centering
    \includegraphics[width=\linewidth]{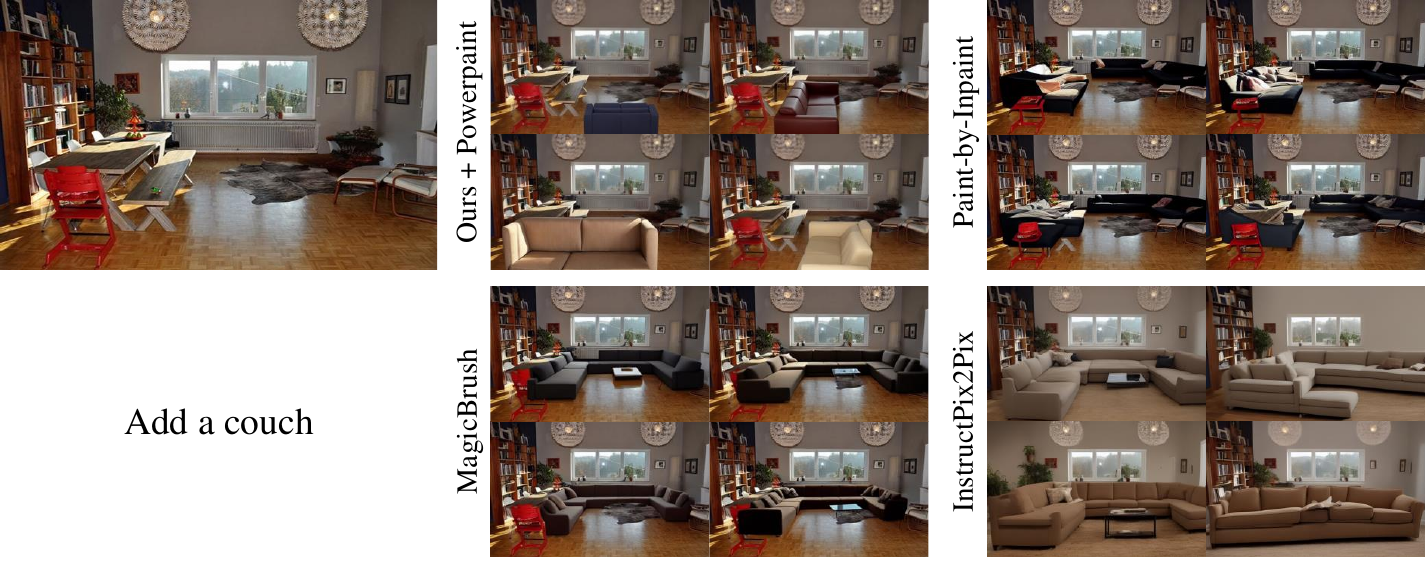}
    \caption{Additional samples of object insertion results in the PIPE test set.}
    \label{fig:appx-quali-pipe0}
\end{figure*}

\begin{figure*}
    \centering
    \includegraphics[width=\linewidth]{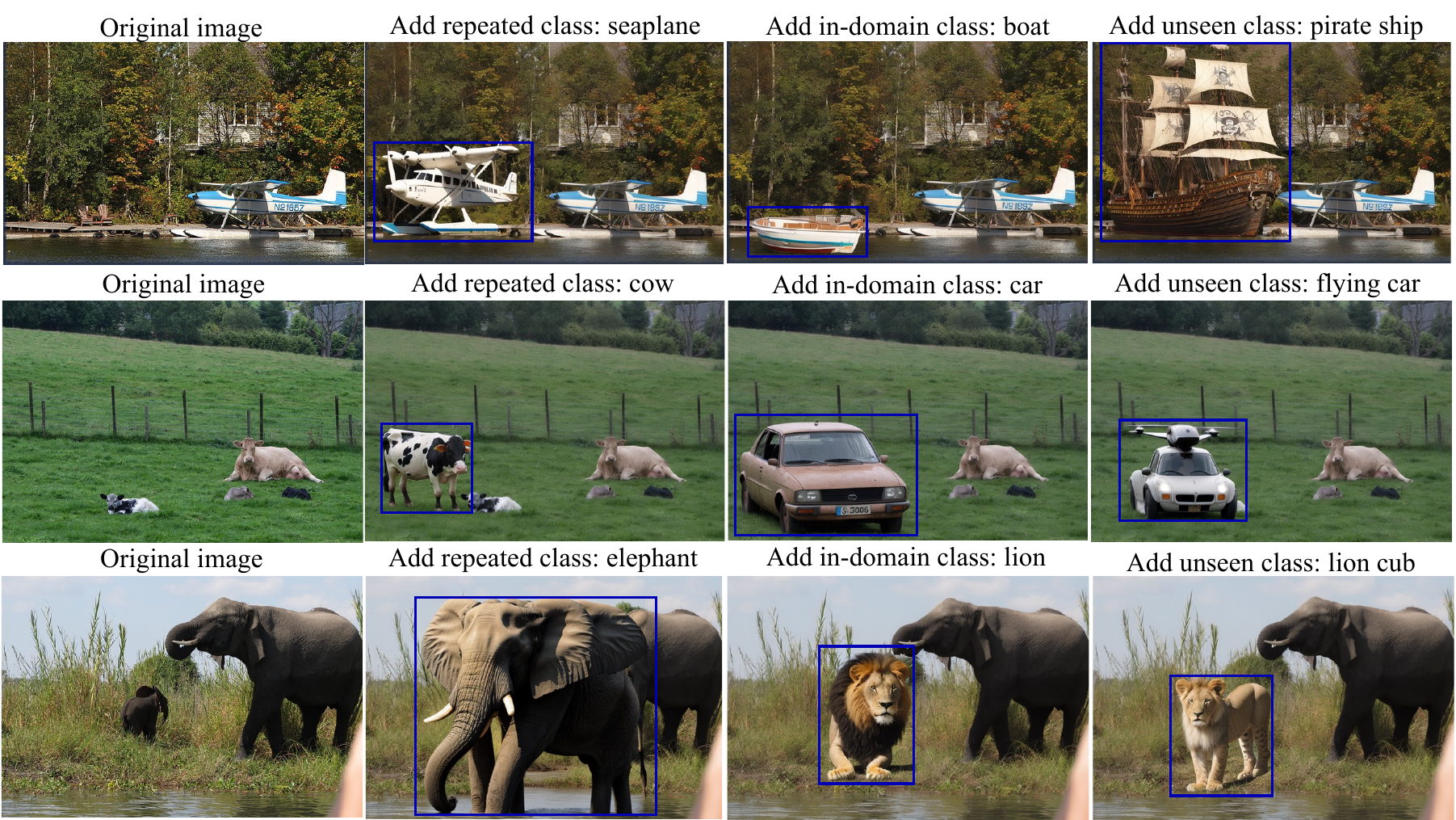}
    \vspace{-0.1cm}
    \caption{Qualitative examples showcasing generalization ability. We ask our model to add one more instance of an already present class (repeated), an instance of a different class (in-domain), and an instance of a class that does not exist in the OPA dataset (unseen).}
    \vspace{-0.3cm}
    \label{fig:appx-generalization1}
\end{figure*}

Additionally, we show qualitative samples highlighting generalization capability in \cref{fig:appx-generalization1}.
As the class is encoded using a CLIP text encoder, we can condition on class labels that are not present in the data and expect a reasonable output if the corresponding CLIP embedding is sufficiently close to existing categories.
We show that we can add another instance of an already present object without our bounding box prediction coinciding with the existing instance, and that we can add instances from other classes in OPA and out-of-domain classes.

